\documentclass{bmvc2k}
\usepackage{enumitem}
\usepackage{cleveref}
\usepackage{algorithm}
\usepackage{algorithmic}
\usepackage{booktabs}
\usepackage{multirow}
\usepackage{amsfonts}

\title{Towards Sharper Object Boundaries in Self-Supervised Depth Estimation
}

\addauthor{Aurélien Cécille}{aurelien.cecille@liris.cnrs.fr}{1,2}
\addauthor{Stefan Duffner}{stefan.duffner@liris.cnrs.fr}{1}
\addauthor{Franck Davoine}{franck.davoine@liris.cnrs.fr}{1}
\addauthor{Rémi Agier}{remi.agier@visualbehavior.ai}{2}
\addauthor{Thibault Neveu}{thibault.neveu@visualbehavior.ai}{2}
\addinstitution{
   LIRIS, UMR5205\\
   INSA Lyon, CNRS, École Centrale de Lyon, Universite Claude Bernard Lyon 1, Université Lumière Lyon 2\\
   69621 Villeurbanne, France
}
\addinstitution{
 Visual Behavior\\ Lyon, France
}

\runninghead{Cécille \etal}{Towards Sharper Object Boundaries}

\def\etal{\emph{et al}\bmvaOneDot}

\begin{document}

\maketitle

\begin{abstract}
   Accurate monocular depth estimation is crucial for 3D scene understanding, but existing methods often blur depth at object boundaries, introducing spurious intermediate 3D points. While achieving sharp edges usually requires very fine-grained supervision, our method produces crisp depth discontinuities using only self-supervision. Specifically, we model per-pixel depth as a mixture distribution, capturing multiple plausible depths and shifting uncertainty from direct regression to the mixture weights. This formulation integrates seamlessly into existing pipelines via variance-aware loss functions and uncertainty propagation. Extensive evaluations on KITTI and VKITTIv2 show that our method achieves up to 35\% higher boundary sharpness and improves point cloud quality compared to state-of-the-art baselines.
\end{abstract}

\section{Introduction}

\begin{figure}[tb]
    \centering
    \includegraphics[width=\textwidth]{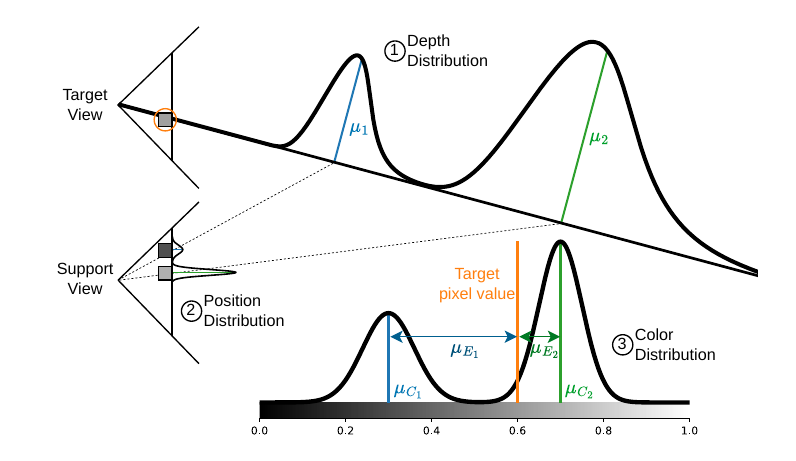}
    \vspace{-20pt}
    \caption{Reprojection pipeline with mixture distributions. \raisebox{.5pt}{\textcircled{\raisebox{-.9pt} {1}}} Predicted depth distribution for each pixel. \raisebox{.5pt}{\textcircled{\raisebox{-.9pt} {2}}}  The depth distribution is projected into the support view, resulting in a distribution of positions. \raisebox{.5pt}{\textcircled{\raisebox{-.9pt} {3}}}  These positions are sampled using bilinear interpolation to obtain a color distribution which is compared to the target color to compute the loss.}
    \label{fig:tsob}
\end{figure}

\begin{figure}[tb]
    \centering
    \includegraphics[width=\textwidth]{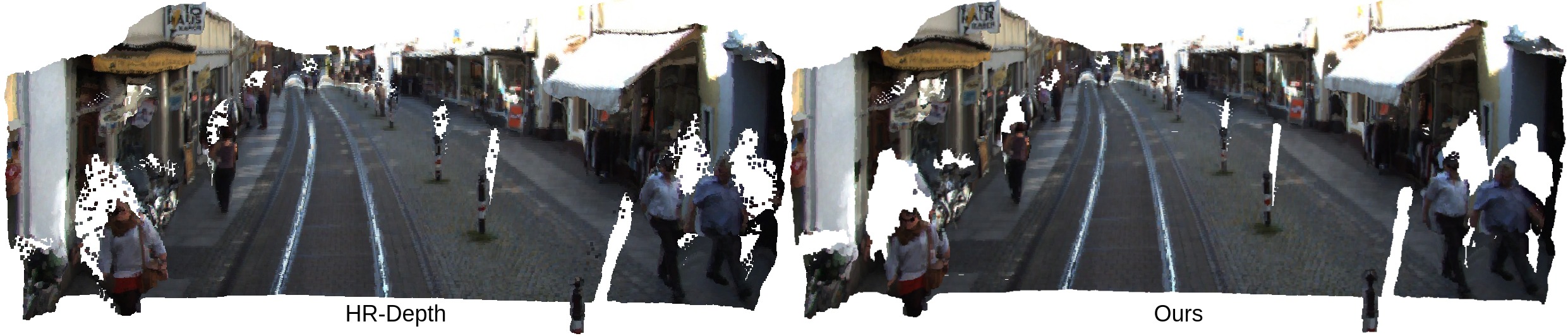}
    \vspace{-20pt}
    \caption{Other methods typically produce floating artifacts at object boundaries, resulting in spurious points between foreground and background. In contrast, our method generates clean point clouds with sharp object boundaries, effectively eliminating these artifacts.}
    \label{fig:pcd}
\end{figure}

Monocular depth estimation is a fundamental problem in computer vision with applications in autonomous driving, robotics and augmented reality.
Recently, self-supervised learning methods have achieved impressive results by using view synthesis as a supervisory signal, but despite these advances, handling depth discontinuities remains challenging. In most scenes, foreground objects occlude the background, creating depth discontinuities at object boundaries. Conventional models assign a single depth value per pixel, but edge uncertainty often causes depth values to be averaged between foreground and background depths, blurring transitions and introducing artifacts in the point cloud (see \Cref{fig:pcd}). To address this, we propose to represent per-pixel depth as a multimodal distribution, explicitly modeling both depths at boundaries, preserving sharp transitions and removing artifacts.

We demonstrate how to integrate this mixture distribution representation into the standard self-supervised depth estimation pipeline by propagating the distributions through the various operations, including reprojection, color interpolation, and loss computation. We address this through careful approximation and uncertainty propagation techniques.

Our main contributions are: \textbf{(1)} A novel mixture distribution representation for self-supervised monocular depth estimation. \textbf{(2)} A principled method for propagating the depth distribution through the reprojection process. \textbf{(3)} An uncertainty-aware loss function that naturally handles multimodal depth distributions. \textbf{(4)} Extensive experiments, including our novel edge entropy measure, confirm our method's strength at object boundaries.

\section{Related Work}
\subsection{Self-Supervised Depth Estimation}

To overcome the reliance on expensive ground truth depth data, Zhou \etal \cite{zhou2017unsupervised} introduced a self-supervised approach that simultaneously learns depth and camera pose from videos using view synthesis as supervision.
This approach has been extended in various directions, including enhancing loss robustness \cite{godard2019digging} and improving model architectures \cite{guizilini20203d,lyu2021hr,zhang_lite-mono_2023,zhao_monovit_2022,wang2024sqldepth, cecille2024groco}. Spencer \etal \cite{spencer2022deconstructing} demonstrated that the choice of backbone significantly impacts performance, providing a systematic comparison using image, point cloud, and boundary metrics. To ensure fair comparison and avoid such pitfalls, we adopt the same backbone as in their study, namely the ConvNeXt-Base \cite{liu2022convnet}.
We also adopt their evaluation metrics since they are highly relevant to our task and address common issues in the literature such as metric saturation and inaccurate labels.

The self-supervised literature focuses primarily on edge position accuracy \cite{chen2023self,lyu2021hr, zhao_monovit_2022}, with some methods exploiting semantic segmentation clues \cite{petrovai2023monodvps, zhou2021self} while others discretize the regression problem by averaging over multiple bins \cite{bhat2021adabins, wang2023planedepth, wang2024sqldepth}.
In contrast, edge sharpness is predominantly addressed in supervised approaches \cite{bochkovskii2024depth,tosi2021smd}, which typically require high-resolution images and fine-grained annotations often sourced from synthetic data. Through our analysis, we discovered that conventional edge sharpness metrics are often confounded with depth regression accuracy. To address this limitation, we propose a novel edge entropy measure that effectively disentangles edge quality assessment from overall depth accuracy.
We also note the approach of Tulsiani \etal \cite{tulsiani2018layer}, which proposes to decompose scenes into multiple depth layers, each learned through view synthesis from different viewpoints.
\subsection{Mixture Models and Density Estimation}
Mixture models are particularly useful for handling multi-modal distributions and have been applied to a variety of tasks in machine learning. Traditionally, the Expectation-Maximization (EM) algorithm \cite{dempster1977maximum} is the method of choice to fit mixture models. Bishop \etal \cite{bishop1994mixture} introduced mixture density networks for modeling conditional probability densities. In computer vision, mixture models have been used to filter outliers in optical flow estimation \cite{ wang2024searaft}, for 3D object tracking \cite{bao2011mixture} and most relevant to our work, Tosi \etal \cite{tosi2021smd} proposed to use a mixture of laplacian distributions for supervised stereo depth estimation.

Mixture models also pose the question of variance and uncertainty estimation in deep learning.
Recent studies on mean-variance estimation \cite{sluijterman2024optimal,skafte2019reliable,seitzer2022pitfalls} highlight the risks of overfitting or underfitting, particularly when using log-likelihood-based approaches.
For depth prediction, Kendall and Gal \cite{kendall2017uncertainties} proposed a Bayesian framework to model both aleatoric and epistemic uncertainty in depth prediction and Poggi \etal \cite{poggi2020uncertainty} provided a comprehensive evaluation of uncertainty estimation techniques in self-supervised monocular depth estimation.
Our work bridges the gap between mixture density models and self-supervised depth estimation, providing a principled way to represent and learn multi-modal depth distributions without ground truth supervision.

\begin{figure}[tb]
    \centering
    \includegraphics[width=\textwidth]{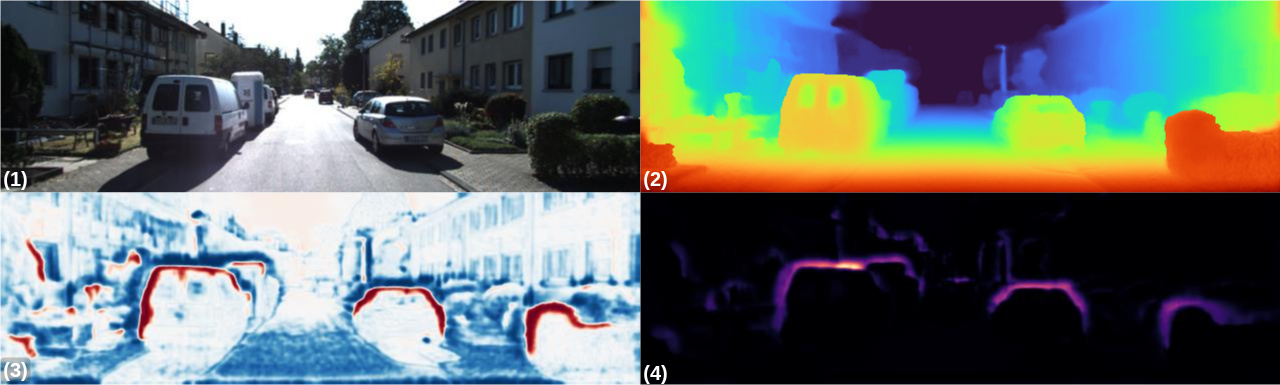}
    \vspace{-15pt}
    \caption{(1) Input Image (2) Predicted Depth (3) Mixture weight: Blue=0 , White=0.5, Red=1 (4) Relative difference between components means. At object boundaries, our model correctly identifies both depth modes and creates sharp discontinuities.
    }
    \label{fig:grid}
\end{figure}

\section{Methodology}
The standard self-supervised depth estimation pipeline is built around view synthesis. Given a pair of nearby frames, we choose one frame as the target and treat the other as source. Two networks are trained together: a depth network that predicts per-pixel depth for the target, and a pose network that predicts the relative camera motion between the two views. Using these predictions and the camera intrinsics, we project pixels from the source into the target to reconstruct how the source should look from the target view. The reconstruction quality provides the learning signal to optimize the networks.

This baseline requires no ground-truth depth and works directly from ordinary video. In the rest of this section, we adopt this framework and introduce a mixture-based representation of depth uncertainty, and show how it is propagated through warping and into the training loss.
\subsection{Distribution Representation}
\label{sec:depth_dist}

For our representation, we work with disparity $d$ (unscaled inverse depth) since it makes distributions more Gaussian and reduces linearization errors in uncertainty propagation \cite{civera2008inverse}.

For each pixel, we assume that the disparity distribution is a mixture of two components, each parameterized by a mean $\mu_k$ and a variance $\sigma_k^2$. There is also a mixing proportion $\alpha$ that controls the relative weight of each component:
\begin{equation}
    p(d) = (1-\alpha)\, p(\, d \, | \, \mu_1, \sigma_1^2) + \alpha \, p(\, d \, | \, \mu_2, \sigma_2^2)
\end{equation}
These five parameters are predicted by a neural network as separate output channels. Following standard practice, we use a sigmoid activation for $\alpha$ and $\mu_k$ and predict the log variance by using an exponential activation.

We use a Gaussian distribution for $p(\, d \, | \, \mu_k, \sigma_k^2)$ since it is a common choice for modeling disparity distributions, allows uncertainty propagation and makes computing the best components analytically tractable.
Note that in this case the depth is not gaussian, since the uncertainty increases with the distance as can be seen in \Cref{fig:tsob}.
\subsection{Distribution Propagation}
To propagate distributions through our pipeline, we model each disparity component as a random variable $D_k \sim \mathcal{N}(\mu_k, \sigma_k^2)$. When applying a differentiable function $f$ to $D_k$, we approximate the result as a normal distribution:
\begin{equation}
    f(D_k) \sim \mathcal{N}(f(\mu_k),\; f'(\mu_k)^2 \sigma_k^2)
\end{equation}
This first-order approximation is accurate when $f$ is approximately linear in the region of interest. In our case, this allows us to propagate uncertainty through the pipeline efficiently and with good accuracy.

The pipeline consists of three main steps shown in \Cref{fig:tsob}: reprojection, color interpolation, and loss computation.
We can define the reprojection function $f_{\text{reproj}}$ that maps the disparity $d$ in the target image to pixel coordinates $(x ,y)$ in the support image.\\
Once we have the position, we use bilinear interpolation to obtain color:
\begin{equation}
    C(x,y) = \begin{bmatrix} 1-x & x \end{bmatrix} \begin{bmatrix} C_{00} & C_{01} \\ C_{10} & C_{11} \end{bmatrix} \begin{bmatrix} 1-y \\ y \end{bmatrix}
\end{equation}
where $C_{ij}$ are the colors of the four nearest pixels.\\
Rather than propagating the uncertainty through these two steps separately, we can define a composite function $s(D_k) = C(f_{\text{reproj}}(D_k))$ that maps directly from disparity to color. Using the chain rule:
\begin{equation}
    s'(D_k) = \nabla C(f_{\text{reproj}}(D_k)) \cdot f'_{\text{reproj}}(D_k)
\end{equation}
where $\nabla C = [\frac{\partial C}{\partial x}, \frac{\partial C}{\partial y}]$ is the gradient of the color interpolation function with respect to position, and $f'_{\text{reproj}}(D_k)$ is the derivative of the reprojection function with respect to disparity.\\
The color distribution is then:
\begin{align}
    \mu_{C_k}\approx       & \ s(\mu_k) = C(f_{\text{reproj}}(\mu_k))                                                                     \\
    \sigma_{C_k}^2 \approx & \ s'(\mu_k)^2 \sigma_k^2 = [\nabla C(f_{\text{reproj}}(\mu_k)) \cdot f'_{\text{reproj}}(\mu_k)]^2 \sigma_k^2
\end{align}
This direct approach handles the propagation of uncertainty through the whole reconstruction pipeline, accounting for how changes in disparity affect both position and ultimately color. It is also very efficient since these computations are done analytically.
\subsection{Loss Distribution and Optimization}
\label{sec:loss_dist}
For the loss computation, we use the combination of the Structural Similarity Index Measure (SSIM) and L1 distance between the target image and the reprojected image. The SSIM is used since it is more robust to small shifts and better at capturing structural information.

Since we have a distribution over the reprojected colors, we can also obtain a distribution over the error values. For each pixel, we define a random variable $E$ representing our error distribution as a weighted sum of SSIM and L1 loss:
\begin{equation}
    E = \lambda_{\text{SSIM}} \cdot (1 - \text{SSIM}) + \lambda_{\text{L1}} \cdot \text{L1}
\end{equation}
For each component $k$ in our mixture model, we compute the mean $\mu_{k, E}$ and variance $\sigma_{k, E}^2$ of this error distribution using the same linear approximation technique described earlier. \\Full derivation is available in the supplementary material.

We then use a competitive training approach where components specialize based on their relative performance. This approach is inspired by moment matching, where we optimize the predicted distribution parameters such that the mean error approaches zero while the variance aligns with the empirically observed error magnitude.

\begin{figure}
    \centering
    \includegraphics[width=0.9\textwidth]{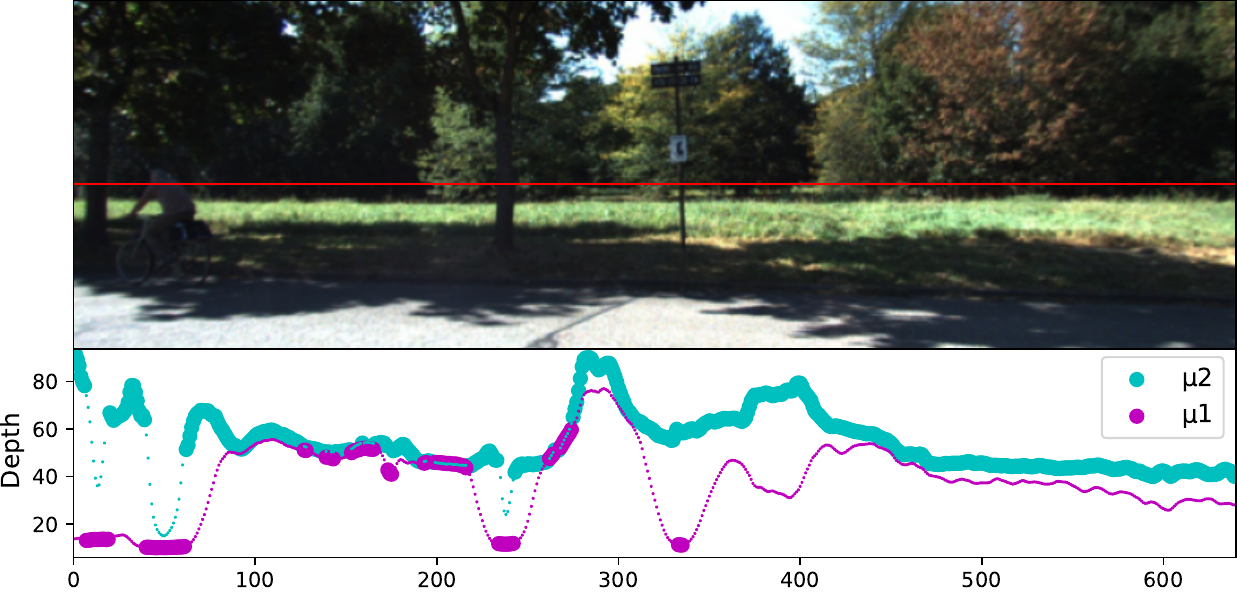}
    \vspace{-10pt}
    \caption{Depth distribution along a single row. Bold points indicate the component selected by the mixture weight. We see that both components complementarily over/underestimate obstacle width smoothly and that it's the selection process that creates the discontinuity.}
    \label{fig:laser_scan}
\end{figure}

In order for our two-component to specialize, we define an intermediate weight variable $w_k$ as the probability that component $k$ has a better (lower) error than the alternative component $1-k$:
\begin{equation}
    w_k = P(E_k < E_{1-k}) = P(E_k - E_{1-k} < 0)
\end{equation}
This last probability can be calculated analytically since both $E_k$ and $E_{1-k}$ are assumed Gaussian and that the difference between two Gaussian random variables is also Gaussian. Specifically, if $E_k \sim \mathcal{N}(\mu_{E_k}, \sigma_{E_k}^2)$ and $E_{1-k} \sim \mathcal{N}(\mu_{E_{1-k}}, \sigma_{E_{1-k}}^2)$, then:
\begin{equation}
    E_k - E_{1-k} \sim \mathcal{N}(\mu_{E_k} - \mu_{E_{1-k}}, \sigma_{E_k}^2 + \sigma_{E_{1-k}}^2)
\end{equation}
Therefore, $w_k$ can be computed using the cumulative distribution function (CDF) of the normal distribution:
\begin{equation}
    w_k = \Phi\left(\frac{\mu_{E_{1-k}} - \mu_{E_k}}{\sqrt{\sigma_{E_{1-k}}^2 + \sigma_{E_k}^2}}\right)
\end{equation}
where $\Phi$ is the CDF of the standard normal distribution.
\\
We then define our loss function with three components:
\begin{align}
    \mathcal{L}_\mu    & = \sum_k \lfloor{w}_k \rceil \mu_{E_k}                    \\
    \mathcal{L}_\sigma & = \sum_k \lfloor{w}_k \rceil (\sigma_{E_k} - \mu_{E_k})^2 \\
    \mathcal{L}_\alpha & = \sum_k w_k \log(\alpha_k)
\end{align}
where $\lfloor{w}_k \rceil$ is the rounded $w_k$, selecting only the most likely component for optimization.

This approach encourages component specialization while maintaining reasonable uncertainty estimates. $\mathcal{L}_\mu$ minimizes the error of the selected component, $\mathcal{L}_\sigma$ aligns the predicted uncertainty with actual error magnitude and $\mathcal{L}_\alpha$ uses cross-entropy to ensure mixture weights reflect component performance. We block gradient propagation through $w_k$ to ensure training stability. We also note that $\mu_{k,E}$ is the default self-supervised loss used in the literature.


\paragraph{Inference.} Likewise, during inference, we select the most likely disparity according to the mixture weights, avoiding the averaging effect caused by its uncertainty.
\begin{equation}
    d^* = \begin{cases}
        \mu_1 & \text{if } \alpha < 0.5 \\
        \mu_2 & \text{otherwise}
    \end{cases}
\end{equation}
\subsection{Edge Sharpness Measure}
\begin{figure}[b]
    \centering
    \includegraphics[width=\linewidth]{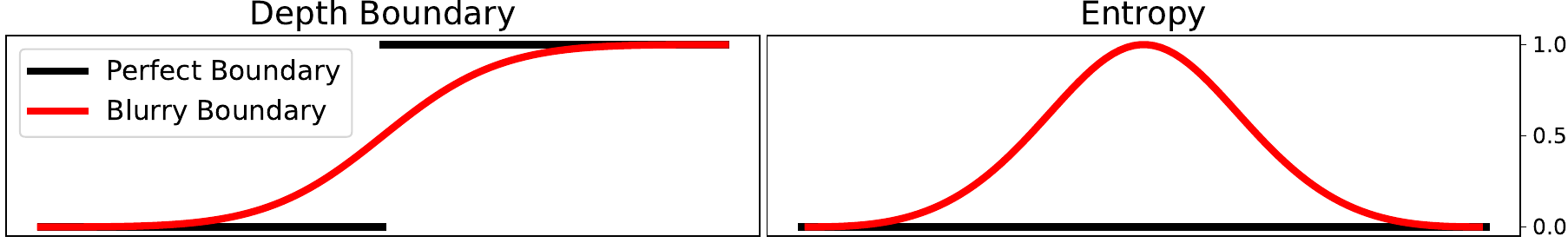}
    \vspace{-20pt}
    \caption{Entropy visualization at object borders: low entropy indicates sharp transitions and high entropy indicates blurry ones.}
    \label{fig:edge_sharpness_graph}
\end{figure}
To quantitatively evaluate the sharpness of depth discontinuities, we introduce a border sharpness measure based on the entropy of edge pixels. This measure captures how cleanly the model represents depth discontinuity, with lower values indicating sharper transitions.
The measure is computed as follows:
\begin{enumerate}[itemsep=0pt, topsep=0pt, parsep=1pt]
    \item Apply a Canny filter to the log depth map to extract edge pixels, forming the set $E$.
    \item For each edge pixel $i \in E$, consider its $3 \times 3$ neighborhood $\mathcal{N}_i$ in the depth map.
    \item Normalize the depth values $d_{ij}$ in $\mathcal{N}_i$ such that $p_{ij} = \frac{d_{ij} - \min(d_{\mathcal{N}_i})}{\max(d_{\mathcal{N}_i}) - \min(d_{\mathcal{N}_i})}$
    \item Compute the Bernoulli entropy for each normalized value:
          \begin{equation}
              H(p_{ij}) = -p_{ij} \log_2(p_{ij}) - (1-p_{ij}) \log_2(1-p_{ij})
          \end{equation}
    \item Average the entropy values within each neighborhood to obtain a local sharpness score for each edge pixel:
          \begin{equation}
              S_i = \frac{1}{|\mathcal{N}_i|} \sum_{j \in \mathcal{N}_i} H(p_{ij})
          \end{equation}
    \item The final sharpness measure is the average over all edge pixels $\mathcal{E}$:
          \begin{equation}
              S = \frac{1}{|\mathcal{E}|} \sum_{i \in \mathcal{E}} S_i
          \end{equation}
\end{enumerate}
As shown in \Cref{fig:edge_sharpness_graph}, this entropy-based approach captures the decisiveness of depth transitions. Values close to 0 or 1 have low entropy (sharp transitions), while values near 0.5 have maximum entropy of 1 (blurry transitions). Using $\text{log}_2$ bounds the entropy between 0 and 1, making the measure interpretable.

\section{Experiments}
\subsection{Datasets}
We train and evaluate our method on the KITTI dataset \cite{geiger2013vision}, a standard benchmark for self-supervised depth estimation featuring stereo pairs and video sequences from urban and highway driving. Following \cite{spencer2022deconstructing}, we use Eigen Zhou splits \cite{eigen2014depth} and report results on the KITTI Eigen Benchmark test set with improved ground truth annotations from \cite{Uhrig2017THREEDV}. VKITTIv2 \cite{gaidon2016virtual, cabon2020vkitti2} is used for its fine-grained synthetic annotations.
We also did some tests on the NYU-Depth v2 dataset \cite{Silberman:ECCV12} to evaluate the generalization of our method to indoor scenes.

\begin{table}
    \centering
    \resizebox{\textwidth}{!}{
        \begin{tabular}{lccccccccc}
            \toprule
            \multirow{2}{*}{Method}                     & \multicolumn{1}{c}{Train} & \multicolumn{4}{c}{Image-based} & \multicolumn{3}{c}{Pointcloud-based}                                                                                                        \\
            \cmidrule(lr){3-6} \cmidrule(lr){7-9}
                                                        & Data                      & MAE$\downarrow$                 & RMSE$\downarrow$                     & AbsRel$\downarrow$ & LogSI$\downarrow$ & Chamfer$\downarrow$ & F-Score$\uparrow$ & IoU$\uparrow$     \\
            \midrule
            SfM-Learner \cite{zhou2017unsupervised}     & M                         & 1.98                            & 4.57                                 & 10.69              & 15.80             & 0.73                & 44.77             & 30.03             \\
            Klodt   \cite{klodt2018supervising}         & M                         & 1.96                            & 4.54                                 & 10.49              & 15.86             & 0.72                & 45.26             & 30.40             \\
            Monodepth2   \cite{godard2019digging}       & M                         & 1.84                            & 4.11                                 & 8.82               & 13.10             & 0.71                & 46.64             & 31.62             \\
            Johnston      \cite{johnston2020self}       & M                         & 1.83                            & \textbf{3.99}                        & 8.85               & 12.89             & 0.71                & 45.72             & 30.78             \\
            HR-Depth      \cite{lyu2021hr}              & M                         & \underline{1.80}                & \underline{4.04}                     & \underline{8.65}   & \textbf{12.75}    & 0.69                & \underline{47.35} & \underline{32.10} \\
            \midrule
            Ours MD2                                    & M                         & 1.81                            & 4.14                                 & 8.92               & 13.35             & \underline{0.68}    & 46.52             & 31.53             \\
            Ours HR-Depth                               & M                         & \textbf{1.76}                   & 4.08                                 & \textbf{8.50}      & \underline{12.86} & \textbf{0.66}       & \textbf{48.46}    & \textbf{33.11}    \\
            \midrule\midrule
            Garg       \cite{garg2016unsupervised}      & S                         & 1.60                            & 3.75                                 & 7.65               & \underline{11.39} & \underline{0.60}    & \underline{53.28} & \underline{37.33} \\
            Monodepth     \cite{godard2017unsupervised} & S                         & 1.61                            & 3.72                                 & 7.73               & 11.57             & 0.64                & 51.25             & 35.45             \\
            SuperDepth      \cite{pillai2019superdepth} & S                         & 1.64                            & 3.77                                 & 7.81               & 11.63             & 0.63                & 52.30             & 36.40             \\
            \midrule
            Ours MD2                                    & S                         & \underline{1.56}                & \underline{3.63}                     & \underline{7.37}   & 11.42             & \underline{0.60}    & 52.32             & 36.75             \\
            Ours HR-Depth                               & S                         & \textbf{1.52}                   & \textbf{3.55}                        & \textbf{7.17}      & \textbf{11.03}    & \textbf{0.58}       & \textbf{53.57}    & \textbf{37.85}    \\
            \midrule\midrule
            Depth-VO-Feat  \cite{zhan2018unsupervised}  & MS                        & 1.63                            & 3.72                                 & 7.70               & 11.64             & \underline{0.62}    & \underline{52.01} & 36.15             \\
            Monodepth2     \cite{godard2019digging}     & MS                        & 1.61                            & 3.62                                 & 7.90               & 10.99             & 0.64                & 50.50             & 34.98             \\
            FeatDepth     \cite{shu2020feature}         & MS                        & 1.60                            & \underline{3.60}                     & 7.80               & 11.01             & 0.65                & 49.99             & 34.51             \\
            CADepth     \cite{yan2021channel}           & MS                        & 1.63                            & 3.60                                 & 8.09               & \underline{10.84} & 0.66                & 49.32             & 34.06             \\
            DiffNet    \cite{zhou2021self}              & MS                        & 1.62                            & 3.63                                 & 7.97               & 10.93             & 0.65                & 49.63             & 34.23             \\
            HR-Depth      \cite{lyu2021hr}              & MS                        & 1.58                            & \textbf{3.56}                        & 7.70               & \textbf{10.68}    & 0.62                & 51.49             & 35.93             \\
            \midrule
            Ours MD2                                    & MS                        & \underline{1.57}                & 3.72                                 & \underline{7.39}   & 11.37             & 0.62                & 51.90             & \underline{36.29} \\
            Ours HR-Depth                               & MS                        & \textbf{1.54}                   & 3.71                                 & \textbf{7.28}      & 11.26             & \textbf{0.61}       & \textbf{52.40}    & \textbf{36.77}    \\
            \bottomrule
        \end{tabular}
    }
    \vspace{-10pt}
    \caption{Comparison of methods on KITTI Eigen Benchmark test set. Results are taken from \cite{spencer2022deconstructing} so that all models use the same backbone and data pipeline.}
    \label{tab:KEB}
\end{table}

\subsection{Implementation Details}

Our encoder uses a ConvNeXt-Base backbone \cite{liu2022convnet} pretrained on ImageNet \cite{deng2009imagenet}. The decoder follows a U-Net-like architecture with 5 output channels: ($\mu_1, \mu_2, \sigma_1, \sigma_2,\alpha$). We use two decoders: Monodepth2 (MD2) \cite{godard2019digging} and HR-Depth \cite{lyu2021hr}.

Each configuration is trained with monocular (M), stereo (S) and both (MS) training data.
We train our model for 30 epochs with a batch size of 8 using the Adam optimizer \cite{kingma2014adam}, an initial learning rate of $10^{-4}$ decreasing to $10^{-5}$ after 15 epochs.
For data augmentation, we apply random horizontal flips and color jitter at a resolution of 640$\times$192.

Our final loss is computed as $\mathcal{L} = \mathcal{L}_{\mu} + \mathcal{L}_{\sigma}  + \lambda_\alpha\mathcal{L}_{\alpha}+\lambda_s \mathcal{L}_s$ with  $\mathcal{L}_s$ the standard smoothness loss \cite{godard2017unsupervised} applied to both mean predictions, $\lambda_s=10^{-3}$ and $\lambda_\alpha=0.1$ to ensure that the gradient of $\mathcal{L}_{\alpha}$ does not dominate the gradient of  $\mathcal{L}_{\mu}$.
We also use the contributions of \cite{godard2019digging}, computing the minimum loss over the different sources and filtering out static pixels.

\begin{figure}[b]
    \centering
    \includegraphics[scale=1]{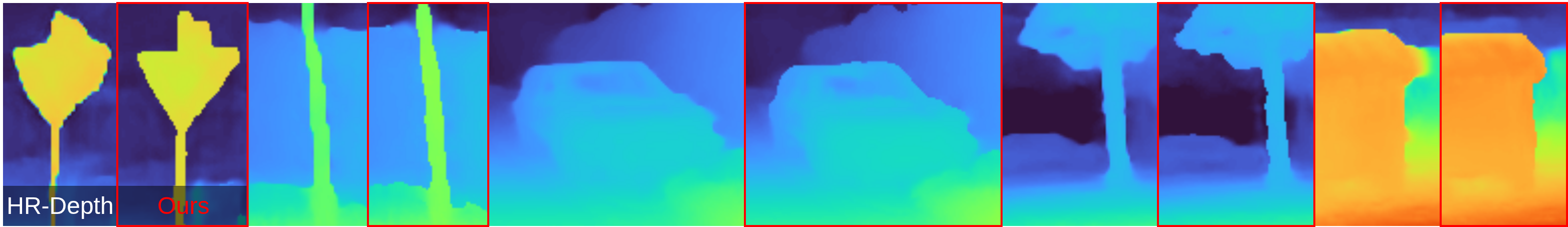}
    \vspace{-20pt}
    \caption{Zoomed-in view of objects with sharp borders. Our method produces noticeably sharper boundary predictions compared to the baseline.}
    \label{fig:zoom}
\end{figure}

\section{Results}
\begin{table}[tb]
    \begin{minipage}[t]{0.48\textwidth}
        \centering
        \begin{tabular}{lccc}
            \toprule
            Method        & M              & MS             & S              \\
            \midrule
            MD2           & 0.499          & 0.520          & 0.509          \\
            Ours MD2      & \textbf{0.350} & \textbf{0.336} & \textbf{0.387} \\
            \midrule
            HR-Depth      & 0.471          & 0.456          & 0.516          \\
            Ours HR-Depth & \textbf{0.373} & \textbf{0.357} & \textbf{0.385} \\
            \bottomrule
        \end{tabular}
        \caption{Edge Sharpness Measure for all methods and data types. Lower values indicate sharper edges.}
        \label{tab:edge_entropy}
    \end{minipage}
    \hfill
    \begin{minipage}[t]{0.48\textwidth}
        \centering
        \begin{tabular}{lccc}
            \toprule
            Method        & M             & MS            & S             \\
            \midrule
            MD2           & 3.53          & 3.70          & 3.68          \\
            Ours MD2      & \textbf{3.31} & \textbf{3.12} & \textbf{3.19} \\
            \midrule
            HR-Depth      & \textbf{2.86} & 2.97          & 3.08          \\
            Ours HR-Depth & 2.88          & \textbf{2.72} & \textbf{2.79} \\
            \bottomrule
        \end{tabular}
        \caption{Edge completeness measure on VKITTIv2, representing the average distance in pixels from predicted edges to labels edges, lower is better. }
        \label{tab:edge_completeness}
    \end{minipage}
\end{table}

\subsection{Quantitative Results}
We compare our method on the KITTI Eigen Benchmark test set in \Cref{tab:KEB} using both image and pointcloud metrics. All methods use the ConvNeXt-Base \cite{liu2022convnet} backbone, and we report results for our method using both MD2 and HR-Depth decoders. We observe modest overall improvements, as expected since edges constitute a small fraction of pixels. We note that when object depth is poorly estimated, sharper boundaries can negatively impact metrics; however, this is mitigated by regions with accurate depth where sharpness is beneficial.

On the edge sharpness measure reported in \Cref{tab:edge_entropy}, which isolates edge quality from overall depth accuracy, our method significantly outperforms baseline methods for both decoders, demonstrating superior performance in predicting discontinuous object boundaries.

We evaluate edge completeness on VKITTIv2 in \Cref{tab:edge_completeness} using the metric from \cite{spencer2022deconstructing}. Our method shows improvements over most baseline methods, indicating that our method is able to more accurately find edges in the depth maps.

\subsection{Qualitative Results}
\Cref{fig:zoom} shows how our method produces sharp depth at object boundaries and \Cref{fig:grid} how it correctly assigns different components to different objects. \Cref{fig:laser_scan} illustrates how the two components tend to work complementarily: one component typically underestimates object width while the other overestimates it. The mixture weight creates the discontinuity at the optimal position by switching between the two components.\\
Side by side depth maps comparisons are also available in the supplementary material.

\subsection{Ablation Studies}
\begin{table}
    \centering
    \resizebox{0.99\textwidth}{!}{
        \begin{tabular}{lccccccccc}
            \toprule
            \multirow{2}{*}{Method} & \multicolumn{4}{c}{Image-based} & \multicolumn{3}{c}{Pointcloud-based}                                                                                                    \\
            \cmidrule(lr){2-5} \cmidrule(lr){6-8}
                                    & MAE$\downarrow$                 & RMSE$\downarrow$                     & AbsRel$\downarrow$ & LogSI$\downarrow$ & Chamfer$\downarrow$ & F-Score$\uparrow$ & IoU$\uparrow$ \\
            \midrule
            Ours MD2                & 1.81                            & 4.14                                 & 8.92               & 13.35             & 0.68                & 46.52             & 31.53         \\
            1 Constant              & 1.82                            & 4.12                                 & 8.96               & 13.29             & 0.69                & 46.08             & 31.18         \\
            2 Constants             & 1.82                            & 4.13                                 & 8.98               & 13.33             & 0.69                & 45.92             & 31.08         \\
            \bottomrule
        \end{tabular}
    }
    \caption{Performance with one variance prediction per pixel (Ours), 1 constant variance for both components (1 Constant), and 1 constant variance for each components (2 Constants).}
    \label{tab:ablation}
\end{table}
In order to understand the contribution of the uncertainty propagation, we perform an ablation study on the KITTI Eigen Benchmark test set with the Monodepth2 architecture. Instead of propagating the uncertainty through the loss, we use constant color variance across the image (with learned scale). In this case we find that the model is not able to separate the two components, underlining the importance of the uncertainty propagation.

Given these results, we then tried to replace the predicted depth variances with constant values. We evaluated two additional configurations: one constant variance per component (2 constants) and a single constant variance for both components (1 constant). Table \ref{tab:ablation} shows that these modifications have minimal impact on performance, suggesting that while variance propagation through the loss is crucial for component separation, the mixture means are the primary drivers of the resulting color variances.

\section{Conclusion and Future Work}
In this paper, we introduced a self-supervised monocular depth estimation method that models each pixel's depth as a mixture of distributions. Our approach achieves competitive accuracy on standard benchmarks and drastically improves depth and pointcloud discontinuities. The proposed distribution propagation technique and mixture-based representation open new possibilities for robust depth estimation and uncertainty modeling in self-supervised learning. These advances have potential applications in related tasks such as optical flow, and can be extended to more complex distributions as well as the integration of temporal and pose uncertainty.
Furthermore, the mixture weights produced by our model provide a rich, interpretable representation of scene structure. These weights highlight object boundaries and regions with distinct depth characteristics, suggesting a possible direction for self-supervised instance segmentation.

\section*{Acknowledgments}

This project was provided with computer and storage resources by GENCI at
IDRIS thanks to the grant 2024-AD011015919 on the supercomputer Jean Zay's H100 partition .

\newpage

\bibliography{references}

\end{document}